\pdfoutput=1

\documentclass[11pt]{article}

\usepackage[preprint]{acl}

\usepackage{times}
\usepackage{latexsym}

\usepackage[T1]{fontenc}

\usepackage[utf8]{inputenc}

\usepackage{microtype}

\usepackage{inconsolata}

\usepackage{graphicx}
\usepackage{booktabs}       %
\usepackage{array}
\usepackage[
  separate-uncertainty = true,
  multi-part-units = repeat
]{siunitx}
\usepackage{multirow}
\usepackage{graphicx}
\usepackage{relsize}
\usepackage{amsmath,amssymb}
\usepackage{tabularx} %
\usepackage{lipsum} %
\usepackage{float}
\usepackage{placeins}
\usepackage{longtable}
\usepackage{enumitem}
\usepackage{stfloats}

\title{EXPERT: An Explainable Image Captioning \\Evaluation Metric with Structured Explanations}

\author{
    \textbf{Hyunjong Kim{\textsuperscript{1}}} \quad
    \textbf{Sangyeop Kim{\textsuperscript{1,2}}} \quad
    \textbf{Jongheon Jeong{\textsuperscript{1}}} \\
    \textbf{Yeongjae Cho{\textsuperscript{1}}} \quad
    \textbf{Sungzoon Cho{\textsuperscript{1$\dagger$}}} \\
    {\textsuperscript{1}}Seoul National University \quad
    {\textsuperscript{2}}Coxwave \\
    \texttt{\{hjkim0811, pdg01117, wwg0713, zxc5932, zoon\}@snu.ac.kr}
}

\begin{document}
\maketitle

\renewcommand{\thefootnote}{$\dagger$}
\footnotetext{Corresponding author.}
\renewcommand{\thefootnote}{\arabic{footnote}}

\begin{abstract}
Recent advances in large language models and vision-language models have led to growing interest in explainable evaluation metrics for image captioning. However, these metrics generate explanations without standardized criteria, and the overall quality of the generated explanations remains unverified. In this paper, we propose EXPERT, a reference-free evaluation metric that provides structured explanations based on three fundamental criteria: fluency, relevance, and descriptiveness. By constructing large-scale datasets of high-quality structured explanations, we develop a two-stage evaluation template to effectively supervise a vision-language model for both scoring and explanation generation. EXPERT achieves state-of-the-art results on benchmark datasets while providing significantly higher-quality explanations than existing metrics, as validated through comprehensive human evaluation. Our code and datasets are available at \url{https://github.com/hjkim811/EXPERT}.
\end{abstract}

\section{Introduction}

Automatic evaluation of image captions is crucial for measuring and improving image captioning models without the significant cost and time required for human evaluation \citep{image_captioning_review}.
Along with the advancement of large language models (LLMs) \citep{jiang2023mistral7b, touvron2023llama2, dubey2024llama, yang2024qwen2, abdin2024phi} and vision-language models (VLMs) \citep{radford2021learning, li2022blip, liu2023visual, liu2024improved}, explainable evaluation metrics have recently drawn increasing attention in image captioning \citep{chan-etal-2023-clair, lee-etal-2024-fleur}. 

Explainable metrics not only provide numerical scores but also textual explanations, enhancing the interpretability and transparency of the evaluation.
However, existing research on explainable metrics has two limitations.
First, the explanations provided by existing metrics are not generated based on standardized criteria or guidelines, which may lead to inconsistency in content and structure.
Second, previous studies lack a thorough evaluation of the quality of generated explanations, leaving the overall quality unverified.

\begin{figure}[t!]
    \centering 
    \includegraphics[width=1.0\linewidth]{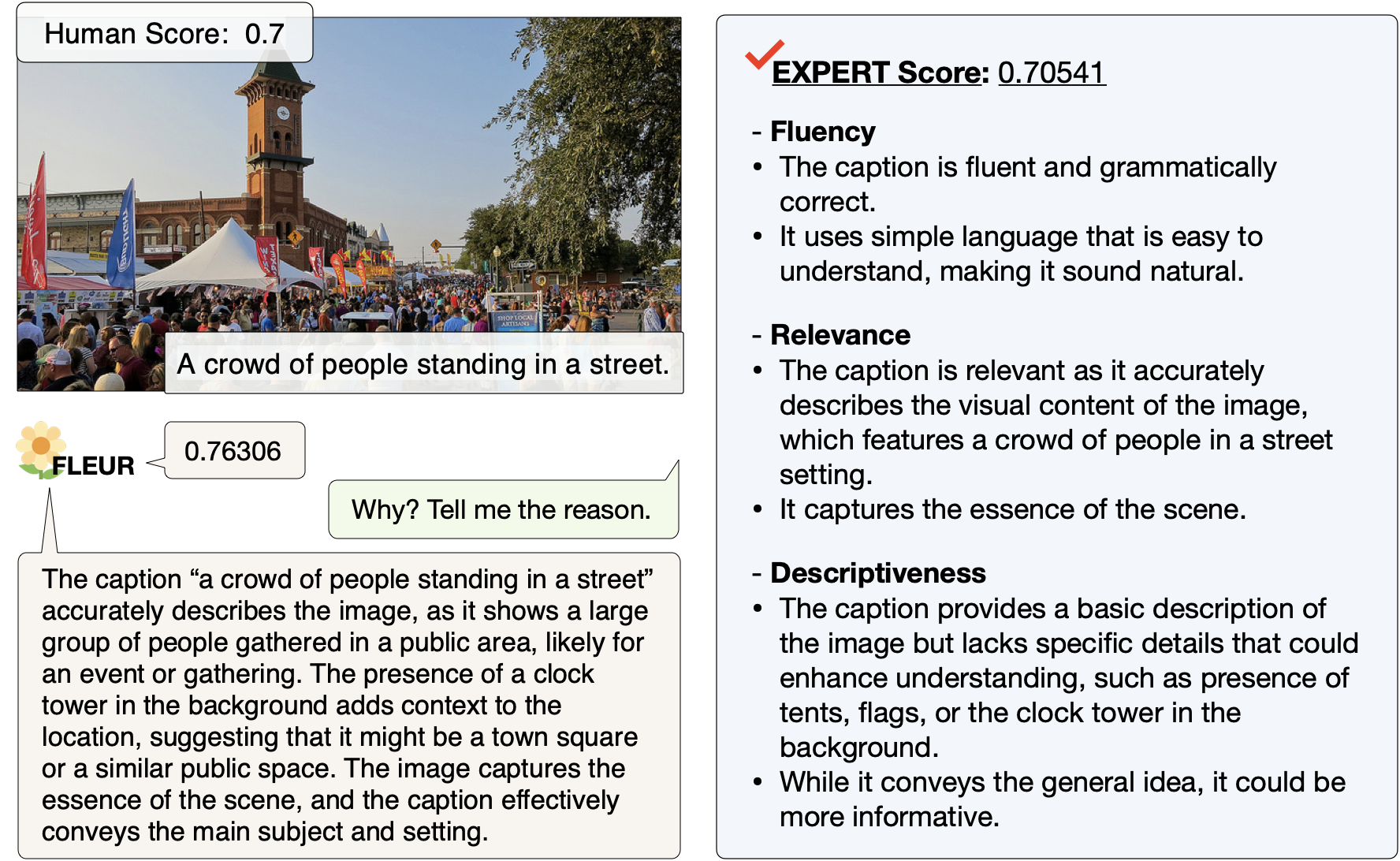}
    \caption{Example of a structured explanation.} 
    \label{fig:intro}
\end{figure}

To address these issues, we propose EXPERT\footnote{\parbox[t]{\linewidth}{An \textbf{EXP}lainable Image Captioning \textbf{E}valuation Metric \\with St\textbf{R}uc\textbf{T}ured Explanations.}}, a VLM-based reference-free evaluation metric for image captioning that provides structured explanations. 
We first construct large-scale datasets containing over 42,000 structured explanations by extending existing human judgment datasets. Each explanation is structured according to three fundamental criteria—fluency, relevance, and descriptiveness—and its quality is validated through human evaluation.
We then train EXPERT using a two-stage evaluation template that effectively guides both scoring and explanation generation.
Unlike existing explainable metrics, EXPERT generates structured, criterion-specific explanations, as illustrated in Figure~\ref{fig:intro}.

Experimental results on benchmark human judgment datasets show that EXPERT establishes new state-of-the-art performance.
Through comprehensive human evaluation, we further demonstrate that EXPERT generates significantly higher-quality explanations than existing explainable metrics.
To the best of our knowledge, we are the first to systematically assess the explanation capabilities of explainable image captioning evaluation metrics.

\section{EXPERT}

\begin{figure*}[t!]
    \centering 
    \includegraphics[width=0.98\linewidth]{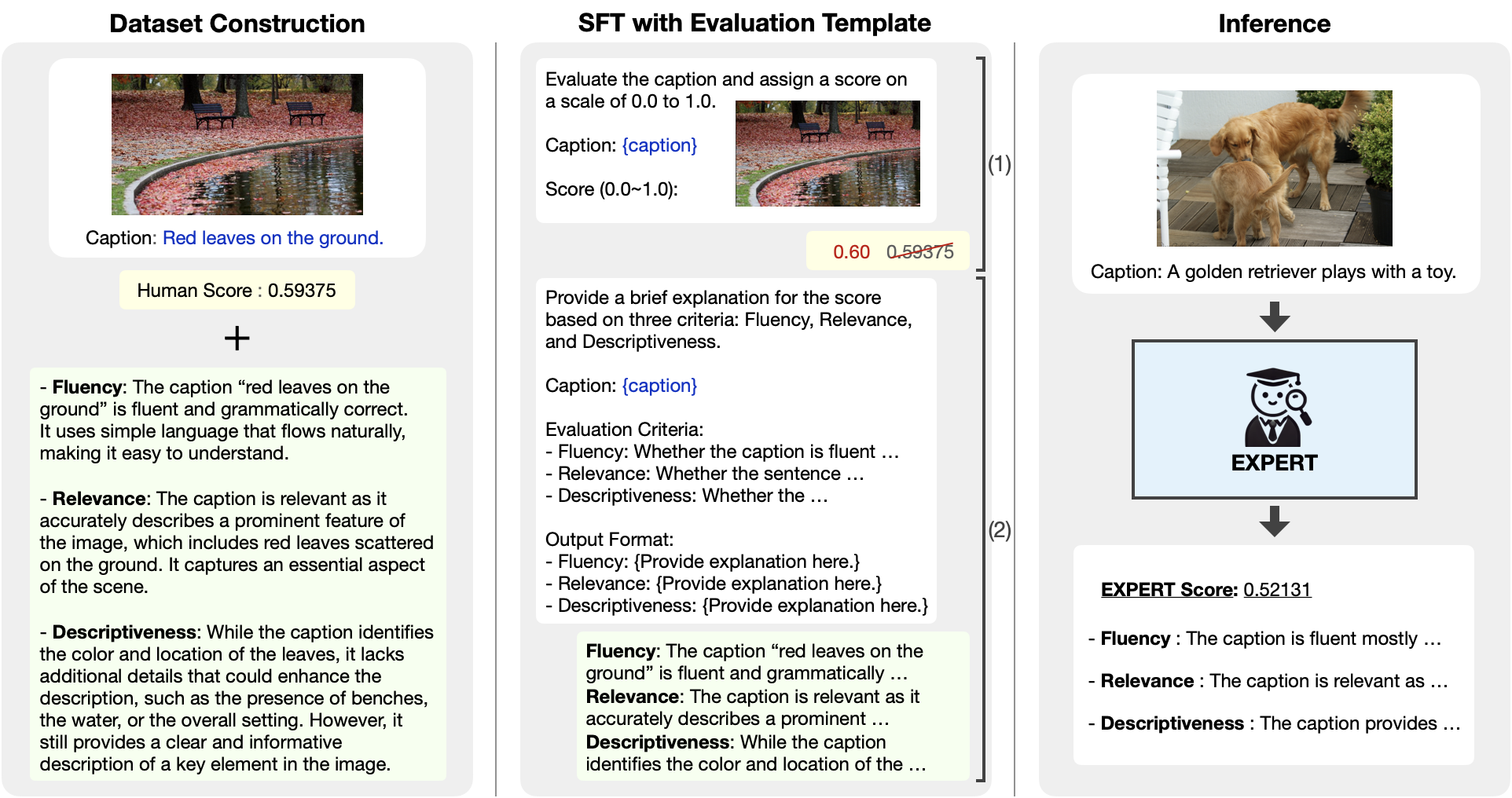}
    \caption{Overall framework of EXPERT.} 
    \label{fig:framework}
\end{figure*}

\subsection{Dataset Construction}
\label{sec:dataset_construction}

\paragraph{Generating Explanations}
Existing image captioning datasets consist of image-caption pairs with the corresponding human judgment scores. However, these datasets lack explanations for the scores, limiting research on the explainability of image captioning evaluation metrics.
To overcome this issue, we construct \textit{Polaris-exp} and \textit{Nebula-exp} datasets by extending the Polaris dataset \citep{wada2024polos} and Nebula dataset \citep{matsuda2024deneb}, respectively. 
We add an \textit{explanation} to each image-caption pair, with each explanation structured according to three dimensions: 
(1) \textbf{Fluency} evaluates whether the caption is fluent, natural, and grammatically correct.
(2) \textbf{Relevance} evaluates whether the sentence correctly describes the visual content and is closely relevant to the image.
(3) \textbf{Descriptiveness} evaluates whether the sentence is a precise, informative caption that describes important details of the image.
These dimensions are the common criteria used for human annotation in the development of recent human judgment datasets~\citep{lee-etal-2021-umic, wada2024polos, matsuda2024deneb}. We generate \mbox{explanations} by \mbox{prompting} GPT-4o \citep{openai2024gpt4ocard} with the prompt presented in Appendix~\ref{sec:generation_prompt} and the corresponding image.
Consequently, Polaris-exp and Nebula-exp contain 16,014 explanations and 26,152 explanations for unique image-caption pairs, respectively.

\paragraph{Dataset Evaluation}
To assess the quality of the generated explanations, we conduct human evaluation on 100 sampled instances from the combined set of Polaris-exp and Nebula-exp.
To ensure balanced representation across different quality levels, we sample uniformly from three score intervals—[0, 0.33), [0.33, 0.66), and [0.66, 1]—with random selection within each interval.
We recruited four annotators native in English to evaluate the explanation quality, particularly based on three criteria:
(1) \textbf{Consistency} measures how logically consistent the explanation is with the given score.
(2) \textbf{Factuality} measures how factually accurate the explanation is in describing the given image-caption pair.
(3) \textbf{Informativeness} measures how much relevant detail and meaningful information the explanation provides to support and justify the given score.
Each annotator was instructed to evaluate all 100 explanations based on these criteria using a 4-point Likert scale.
Scoring guidelines and the evaluation form are provided in Appendix~\ref{sec:human_eval}.

The evaluation results in Table~\ref{tab:dataset_eval} indicate that the explanations in Polaris-exp and Nebula-exp exhibit high quality while maintaining strong alignment with the original dataset components.

\begin{table}[t]
  \small
  \centering
    \begin{tabular}{lrr}
      \toprule
      \textbf{Criteria}     & \textbf{Average} & \textbf{Std. Dev.} \\
      \midrule
      Consistency         & 3.72             & 0.52              \\
      Factuality          & 3.84             & 0.39              \\
      Informativeness     & 3.72             & 0.45              \\
      \bottomrule
    \end{tabular}
  \caption{Human evaluation results for generated explanations in Polaris-exp and Nebula-exp. Std. Dev. represents the overall standard deviation across all ratings for each criterion.}
  \label{tab:dataset_eval}
\end{table}

\subsection{Two-stage Evaluation Template}
Based on the constructed Polaris-exp and Nebula-exp datasets, we design a \textit{two-stage evaluation template} to effectively supervise a general-purpose VLM with human preferences and high-quality structured explanations.
As illustrated in Figure~\ref{fig:framework}, the template follows a scoring-explanation order, which has been shown to be effective in previous studies \citep{chan-etal-2023-clair, lee-etal-2024-fleur}.

The first \textit{scoring stage} contains a query to assign a score to the given image-caption pair, followed by a response which is a human score from the constructed datasets.
At this stage, we apply \textit{score binning}, an operation that rounds scores to the nearest multiple of a specified bin size, for simplified numerical representation (see Appendix~\ref{sec:score_binning} for details). We use a bin size of 0.10 for our experiments.

The second \textit{explanation stage} contains a query to provide a brief explanation for the score based on three criteria—\textit{fluency}, \textit{relevance}, and \textit{descriptiveness}—along with a description for each criterion and a predefined output format.
The following response is the corresponding explanation from the constructed datasets.
For each criterion, we adopt the same descriptions used in Section~\ref{sec:dataset_construction} to maintain consistency.
The full text of the two-stage evaluation template is available in Appendix~\ref{sec:eval_template}.

\subsection{Supervised Fine-tuning}
We use LLaVA-1.5 (13B) \citep{liu2024improved} as our base model for supervised fine-tuning (SFT). The combined training splits of Polaris-exp and Nebula-exp are converted into two-stage evaluation templates for training. Since Polaris contains duplicate image-caption pairs with scores from different annotators, we group these into single instances by averaging the human scores. Additionally, as \mbox{Polaris} and Nebula share overlapping image-caption pairs across the two datasets, we merge these cross-dataset duplicates into single instances by averaging the human scores. Implementation details for SFT are provided in Appendix~\ref{sec:sft_details}.

\subsection{Inference}
Given an image-caption pair, we sequentially prompt the trained model with the queries of the two-stage evaluation template.
We use greedy decoding to ensure deterministic and reproducible results, and apply score smoothing \citep{lee-etal-2024-fleur} in the scoring stage to obtain more detailed scores.
Score smoothing involves computing the probability $p(i,j)$ for each digit $i$ ($0 \le i \le 9$) being generated at the $j$-th decimal place ($j=1,2$), which is then used to produce the final score $s$:
\begin{equation*}
s=\sum_{j=1}^2 10^{-j}\sum_{i=0}^9i\times p(i,j).
\end{equation*}

\section{Experiments}

\subsection{Evaluation Settings}
We evaluate EXPERT on benchmark human judgment datasets: Flickr8k-EX, Flickr8k-CF \citep{hodosh2013framing}, COMPOSITE \citep{aditya2015images}, Polaris \citep{wada2024polos}, Nebula \citep{matsuda2024deneb}, and Pascal-50S \citep{vedantam2015cider}.
For Polaris and Nebula, we use the test split to ensure no overlap with the training data.
Following previous studies, we use  Kendall-B ($\tau_{b}$) for Flickr8k-CF, accuracy for Pascal-50S, and Kendall-C ($\tau_{c}$) for the other datasets. 
Further details of these datasets are provided in Appendix~\ref{sec:benchmarks}.

We compare the performance of EXPERT with various reference-free and reference-based evaluation metrics, including recently proposed methods such as FLEUR \citep{lee-etal-2024-fleur}, HICE-S \citep{zeng2024hicescore}, and DENEB \citep{matsuda2024deneb}. 
Detailed descriptions of the baseline metrics are provided in Appendix~\ref{sec:baselines}.
We also include a comparison with an LLM-as-a-judge approach by directly prompting GPT-4o \citep{openai2024gpt4ocard}, a representative proprietary LLM.
Experimental details for GPT-4o are provided in Appendix~\ref{sec:gpt_details}.

\renewcommand{\arraystretch}{1.05}
\begin{table*}[ht!]
\centering
\resizebox{\textwidth}{!}{%
{\small
\begin{tabular}{clcccccccccc} %
\toprule
  \multirow{2.5}{*}{\textbf{Type}} & \multirow{2.5}{*}{\textbf{Metric}} & \multicolumn{2}{c}{\textbf{Flickr8k}}  & \textbf{COM} & \textbf{Polaris} & \textbf{Nebula} & \multicolumn{5}{c}{\textbf{Pascal-50S} (Accuracy $\uparrow$ )} \\
\cmidrule(lr){3-4}
\cmidrule(lr){5-5}
\cmidrule(lr){6-6}
\cmidrule(lr){7-7}
\cmidrule(lr){8-12}
&& EX ($\tau_c\uparrow$ ) & CF ($\tau_b\uparrow$ ) & ($\tau_c\uparrow$ ) & ($\tau_c\uparrow$ ) & ($\tau_c\uparrow$ ) & HC & HI & HM & MM & Avg  \\
\midrule
\multirow{15}{*}{\shortstack{Reference-based}} 
& BLEU-4 & 30.8 & 16.9 & 30.6 & 46.3 & 40.4 & 53.0 & 92.4 & 86.7 & 59.4 & 72.9 \\
& ROUGE-L & 32.3 & 19.9 & 32.4 & 46.3 & 42.6 & 51.5 & 94.5 & 92.5 & 57.7 & 74.1 \\
& METEOR & 41.8 & 22.2 & 38.9 & 51.2 & 46.8 & 56.7 & 97.6 & 94.2 & 63.4 & 78.0 \\
& CIDEr & 43.9 & 24.6 & 37.7 & 52.1 & 48.1 & 53.0 & 98.0 & 91.5 & 64.5 & 76.8 \\
& SPICE & 44.9 & 24.4 & 40.3 & 51.0 & 44.0 & 52.6 & 93.9 & 83.6 & 48.1 & 69.6 \\
& BERTScore & 39.2 & 22.8 & 30.1 & 51.6 & 47.0 & 65.4 & 96.2 & 93.3 & 61.4 & 79.1 \\
& CLAIR & 48.3 & 34.4 & \underline{61.0} & \textit{53.3} & 52.7 & 52.4 & 99.5 & 89.8 & 73.0 & 78.7 \\
\cmidrule{2-12}
& TIGEr & 49.3 & -- & 45.4 & -- & -- & 56.0 & \textbf{99.8} & 92.8 & 74.2 & 80.7 \\
& ViLBERTScore-F & 50.1 & -- & 52.4 & -- & -- & 49.9 & 99.6 & 93.1 & 75.8 & 79.6 \\
& RefCLIPScore & 53.0 & 36.4 & 55.4 & 52.3 & 46.9 & 64.5 & 99.6 & 95.4 & 72.8 & 83.1 \\
& RefPAC-S & 55.9 & 37.6 & 57.3 & 56.0 & 50.6 & 67.7 & 99.6 & 96.0 & 75.6 & 84.7 \\
& Polos & 56.4 & 37.8 & 57.6 & 57.8 & 53.9 & 70.0 & 99.6 & 97.4 & \underline{79.0} & 86.5 \\
& RefFLEUR & 51.9 & \textbf{38.8} & \textbf{64.2} & \textbf{\textit{58.8}} & \textbf{\textit{54.4}} & 68.0 & \textbf{99.8} & \textbf{98.0} & 76.1 & 85.5 \\
& RefHICE-S & \textbf{57.7} & \underline{38.2} & 58.7 & -- & -- & \underline{71.4} & \underline{99.7} & \underline{97.7} & \textbf{79.7} & \textbf{87.3} \\
& DENEB & \underline{56.5} & 38.0 & 57.9 & \underline{\textit{58.2}} & \underline{54.1} & \textbf{74.4} & \textbf{99.8} & 97.3 & 76.5 & \underline{87.0} \\
\midrule
\multirow{7}{*}{\shortstack{Reference-free}} 
& CLIPScore & 51.2 & 34.4 & 53.8 & 52.3 & 46.9 & 56.5 & 99.3 & 96.4 & 70.4 & 80.7 \\
& PAC-S & 54.3 & 36.0 & 55.7 & 52.5 & 47.2 & 60.6 & 99.3 & 96.9 & 72.9 & 82.4 \\
& InfoMetIC+\footnotemark{} & 55.5 & 36.6 & 59.3 & -- & -- & -- & -- & -- & -- & --  \\
& FLEUR & 53.0 & \underline{38.6} & \underline{63.5} & \underline{\textit{58.3}} & \underline{\textit{51.7}} & 61.3 & \underline{99.7} & \underline{97.6} & 74.2 & 83.2 \\
& BRIDGE & 54.8 & 36.1 & 55.0 & -- & -- & 59.4 & 99.4 & 97.5 & 74.0 & 82.6 \\
& HICE-S & \underline{56.4} & 37.2 & 57.9 & -- & -- & \textbf{68.6} & \underline{99.7} & 96.5 & \textbf{79.5} & \textbf{86.1} \\
& EXPERT & \textbf{56.7} & \textbf{39.3} & \textbf{65.0} & \textbf{61.1} & \textbf{54.9} & \underline{62.8} & \textbf{99.8} & \textbf{97.8} & \underline{78.4} & \underline{84.7} \\
\midrule
\multirow{1}{*}{\shortstack{LLM-as-a-judge}} 
& GPT-4o & \textit{54.3} & \textit{39.3} & \textit{65.9} & \textit{58.2} & \textit{54.3} & \textit{60.8} & \textit{99.7} & \textit{97.3} & \textit{73.7} & \textit{82.9} \\
\bottomrule
\end{tabular}
}}
\caption{Comparison of correlation coefficients and accuracy on various human judgment datasets. \textbf{Bold} and \underline{underlined} values indicate the best and second-best results for each type, respectively. \textit{Italicized} values indicate results reproduced or implemented in this work, while the others are reported from previous works. The `--' symbol indicates unreproducible values due to non-executable code or unavailable data. In the reference-based section, the upper block contains metrics that rely solely on reference captions, while the lower block contains metrics that use both the image and the reference caption.}
\label{tab:main_results}
\end{table*}

\subsection{Correlation with Human Judgments}

The evaluation results of our experiments are shown in Table~\ref{tab:main_results}.
EXPERT achieves state-of-the-art results on all benchmark datasets among reference-free metrics, except for Pascal-50S. On Pascal-50S HC, MM, and Avg, HICE-S attains higher accuracy, while EXPERT still achieves the second-best results.
Notably, EXPERT even outperforms existing reference-based metrics on Flickr8k-CF, COMPOSITE, Polaris, and Nebula.

When compared to GPT-4o, EXPERT achieves equal or superior results across all datasets except for COMPOSITE.
GPT-4o exhibits relatively lower performance despite its substantially larger model size, which we attribute to two main factors: tokenizer differences and limited access to token probabilities. A detailed discussion of these issues is provided in Appendix~\ref{sec:gpt_results}.

These results demonstrate the effectiveness of SFT with the proposed evaluation template.
The superior performance of EXPERT on datasets other than Polaris and Nebula suggests that human preferences across different datasets and annotators exhibit a certain degree of consistency, making it a robust and generalizable evaluation metric.

\footnotetext{We do not include the Pascal-50S results of InfoMetIC+ as \citet{lee-etal-2024-fleur} cast doubt on the experimental settings. Despite our efforts to reproduce the results, the necessary resources for reproduction were inaccessible.}

\subsection{Evaluation of Explanation Quality}
\label{sec:eval_explanation}

\begin{figure*}[!t]
    \centering 
    \includegraphics[width=1.0\linewidth]{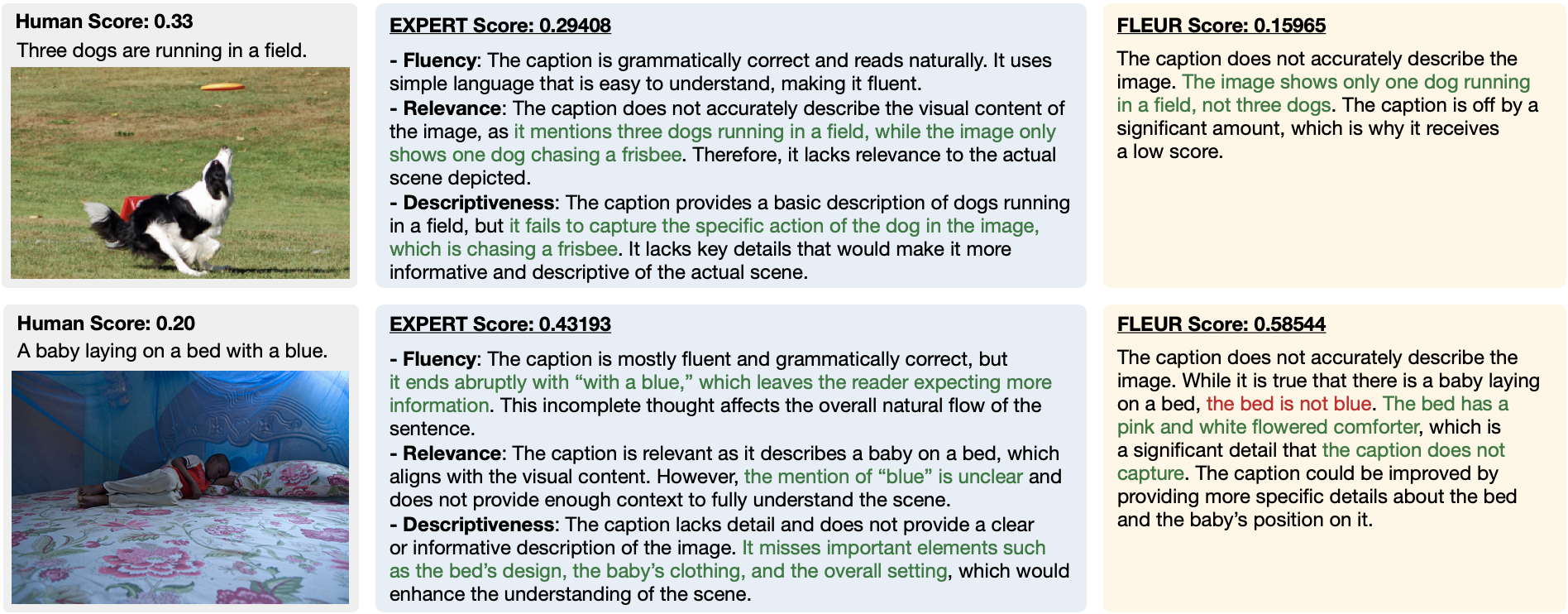}
    \caption{Examples of explanations generated by EXPERT and FLEUR. Green text indicates valid justifications for the given score, while red text indicates incorrect or misleading explanations.} 
    \label{fig:main_examples}
\end{figure*}

\paragraph{Qualitative Example}
Figure~\ref{fig:main_examples} presents examples of explanations generated by EXPERT and FLEUR. In the first example, both metrics correctly identify that the image contains only one dog, not three. However, while FLEUR overlooks the omission of the frisbee in the caption, EXPERT explicitly points out the lack of detail regarding the dog's action—chasing a frisbee.
The second example demonstrates a caption with an incomplete sentence structure. While FLEUR misinterprets the caption as mentioning ``a blue bed'', EXPERT accurately recognizes the grammatical incompleteness and provides appropriate explanations.
Additional examples are provided in Appendix~\ref{sec:additional_examples}.

\paragraph{Human Evaluation}
Through human evaluation, we compare the overall quality of explanations generated by EXPERT and FLEUR, the only two metrics that are both explainable and reference-free.
We introduce an additional baseline, $\text{EXPERT}_{\text{w/o SFT}}$, which prompts LLaVA-1.5 (13B) with the queries of the two-stage evaluation template without SFT.
For each metric, we evaluate explanations generated for the same 100 image-caption pairs sampled from Flickr8k-EX.
To ensure diversity across different quality levels, we sample uniformly for each human score (1, 2, 3, and 4), with random selection within each score.
Four native English-speaking annotators each evaluated the explanations for all three metrics across all 100 samples, based on the same criteria and scoring guidelines used in Section~\ref{sec:dataset_construction}.
Unlike the previous evaluation, the given scores in this evaluation are also generated by the metrics themselves.
The evaluation form is provided in Appendix~\ref{sec:eval_form}.

\begin{figure}[!t]
    \centering 
    \includegraphics[width=1.0\linewidth]{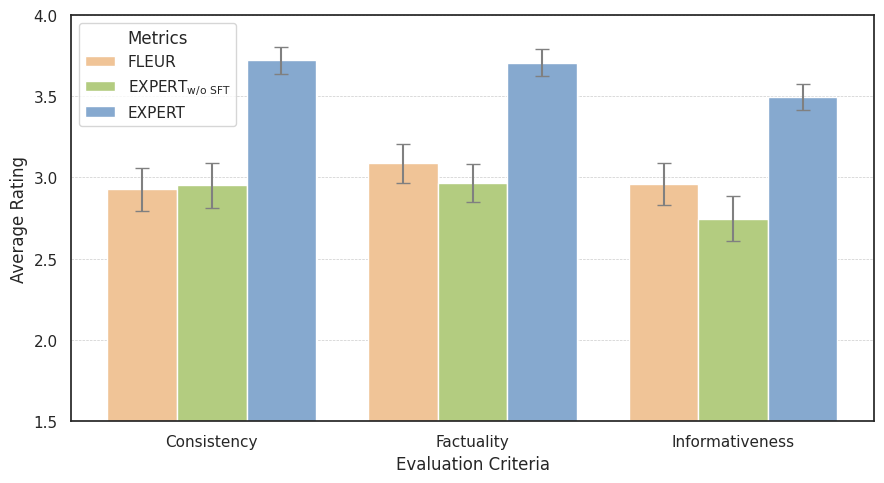}
    \caption{Human evaluation results on explanation quality. Error bars represent the 95\% confidence intervals. Differences between EXPERT and the other metrics are statistically significant for all criteria at the 0.01 significance level.} 
    \label{fig:human_eval}
\end{figure}

The evaluation results are illustrated in Figure~\ref{fig:human_eval}.
EXPERT achieves the highest average rating across all criteria by a large margin, demonstrating that it provides the most logically consistent, factually accurate, and informative explanations for the scores.
$\text{EXPERT}_{\text{w/o SFT}}$, however, demonstrates comparable or slightly lower performance than FLEUR.
These results suggest that while the use of standardized and detailed criteria alone is insufficient to enhance explanation quality, it can lead to significant improvements when combined with supervision using high-quality explanations.

\section{Conclusion}

In this work, we propose EXPERT, a state-of-the-art reference-free evaluation metric that provides structured explanations based on standardized criteria. Along with our novel explanation datasets and analysis of explanation quality, we hope this work offers a valuable contribution to advancing the explainability of image captioning evaluation.

\section*{Limitations}

\paragraph{Error Analysis}
Although EXPERT generally performs well in scoring and generating explanations, it sometimes fails to provide accurate assessments or explanations. We analyzed the 100 samples with the largest absolute difference between the EXPERT score and human score. The errors could be grouped into six main categories, where the most common type of error was the \textit{overpenalization of captions lacking details}. Further details of the error analysis are provided in Appendix~\ref{sec:error_analysis}.

\paragraph{Inference Time} 
The inference times for EXPERT and FLEUR are shown in Table~\ref{tab:inference_time} of Appendix~\ref{sec:inference_time}.
Since both metrics adopt a scoring-explanation order, it is possible to obtain only the scores for faster inference, represented as $\text{EXPERT}_{\text{score}}$ and $\text{FLEUR}_{\text{score}}$.
As the inference time of VLMs heavily depends on the number of output tokens, generating both the score and textual explanations requires significantly more time than generating only the score.
Additionally, EXPERT takes longer than FLEUR due to its longer explanations.
The issue of inference time in explainable metrics should be resolved for practical usability.

\section*{Ethics Statement}

\paragraph{Human Evaluation}
In this work, we conduct two types of human evaluation: evaluation of Polaris-exp and Nebula-exp datasets, and evaluation of the explanations generated by explainable metrics. 
The same group of four annotators participated in both evaluations and were compensated at a rate of USD \$15 per hour.
Our human evaluation qualifies as exempt from IRB approval according to U.S. federal regulation 45 CFR 46.104(d)(2)(i), as we did not collect any personally identifiable information and recorded the data in such a manner that the identity of participants could not be readily ascertained.

\section*{Acknowledgments}
This work was supported by the BK21 FOUR Program (Education and Research Center for Industrial Innovation Analytics) funded by the Ministry of Education, Korea (No. 4120240214912) and the National Research Foundation of Korea (NRF) grant funded by the Korea government (MSIT) (No. 2021R1A2C2093785).

\bibliography{anthology,custom}

\clearpage

\appendix

\section{Prompt for Generating Explanations}
\label{sec:generation_prompt}

\noindent\texttt{Your task is to evaluate the given caption based on three criteria: Fluency, Relevance, and Descriptiveness.
Provide a brief explanation for each criterion without assigning a numerical score.
\newline
\newline
Caption: \textbf{\{caption\}}
\newline
\newline
Criterion 1: Fluency
\newline
- Whether the caption is fluent, natural, and grammatically correct.
\newline
\newline
Criterion 2: Relevance
\newline
- Whether the sentence correctly describes the visual content and is closely relevant to the image.
\newline
\newline
Criterion 3: Descriptiveness
\newline
- Whether the sentence is a precise, informative caption that describes important details of the image.
\newline
- If the caption includes the necessary key details to provide a clear and informative description of the image, it should be considered descriptive enough. In this case, do not say it lacks detail or it could be more informative.
\newline
\newline
Output format:
\newline
Fluency: \{Provide explanation here.\}
\newline
Relevance: \{Provide explanation here.\}
\newline
Descriptiveness: \{Provide explanation here.\}
\newline
} 

In preliminary experiments, we found that GPT-4o is overly rigorous in assessing descriptiveness, generating critical feedback such as ``it could be more informative'' even for captions that received high human scores. To address this challenge, we enhanced the description for descriptiveness with additional specifications.

We also tried providing the human score in the prompt, but this often led to biased explanations, such as generating negative feedback across all three criteria for captions with low human scores, even when some criteria demonstrated high quality.
Therefore, we excluded human scores from the prompt to ensure objective explanations.

\newpage

\section{Two-stage Evaluation Template}
\label{sec:eval_template}

\subsection{Scoring Stage: Query}
\noindent\texttt{Evaluate the caption and assign a score on a scale of 0.0 to 1.0. 
\newline
\newline
Caption: \textbf{\{caption\}}
\newline
\newline
Score (0.0\textasciitilde1.0):
\newline
}

\subsection{Scoring Stage: Response}
\noindent\texttt{\textbf{\{score\}}\newline}

\subsection{Explanation Stage: Query}
\noindent\texttt{Provide a brief explanation for the score based on three criteria: Fluency, Relevance, and Descriptiveness.
\newline
\newline
Caption: \textbf{\{caption\}}
\newline
\newline
Evaluation Criteria:
\newline
- Fluency: Whether the caption is fluent, natural, and grammatically correct.
\newline
- Relevance: Whether the sentence correctly describes the visual content and is closely relevant to the image.
\newline
- Descriptiveness: Whether the sentence is a precise, informative caption that describes important details of the image.
\newline
\newline
Output Format:
\newline
Fluency: \{Provide explanation here.\}
\newline
Relevance: \{Provide explanation here.\}
\newline
Descriptiveness: \{Provide explanation here.\}
\newline
}

\subsection{Explanation Stage: Response}
\noindent\texttt{Fluency: \textbf{\{explanation for fluency\}}
\newline
Relevance: \textbf{\{explanation for relevance\}}
\newline
Descriptiveness: \textbf{\{explanation for \\descriptiveness\}}
}

\clearpage
\onecolumn

\section{Human Evaluation}
\label{sec:human_eval}

\subsection{Evaluation Criteria and Scoring Guidelines}
\label{sec:human_eval_criteria}

Table~\ref{tab:human_eval_criteria} shows the evaluation criteria and detailed scoring guidelines for each criterion.
A 4-point scale was chosen over a 5-point scale to avoid neutral responses and encourage more decisive judgments.
For informativeness, we added a `Disagreement Option' to address cases where the score is severely misaligned with the image-caption pair, making it difficult to evaluate the explanation meaningfully.
We intended to exclude these cases from result aggregation, but this option was not selected by any evaluator for any data sample.

{\renewcommand{\arraystretch}{1.1}
    \begin{table*}[ht!] 
    \small
    \begin{center}
    \setlength{\tabcolsep}{3pt}
    \begin{tabularx}{\linewidth}{X}
        \toprule
        
        \textbf{Consistency} \\
        How logically consistent is the explanation with the score? \\
        Consider only the explanation text and the numerical score, without referencing the image. \\ \\
        
        Scoring Guidelines: \\
        
        - 1: The nuance and tone of the explanation are \textbf{completely misaligned} with the score, offering contradictory reasoning that fails to justify the score in any meaningful way. \\
        - 2: The nuance and tone of the explanation are \textbf{partially aligned} with the score, with noticeable mismatches or gaps in logic that make the connection unclear. \\
        - 3: The nuance and tone of the explanation are \textbf{mostly aligned} with the score, but minor inconsistencies or subtle mismatches slightly reduce its overall coherence. \\
        - 4: The nuance and tone of the explanation are \textbf{fully aligned} with the score, leaving no doubt about the connection between the explanation and the score. \\

        \midrule
        
        \textbf{Factuality} \\
        How factually accurate is the explanation in describing the image-caption pair? \\ \\
        
        Scoring Guidelines: \\
        
        - 1: The explanation contains \textbf{critical factual errors} or describes something \textbf{entirely unrelated or contradictory} to the image-caption pair. \\
        - 2: The explanation is \textbf{partially factual}, with some accurate elements but significant incorrect details. \\
        - 3: The explanation is \textbf{mostly factual}, with only minor factual inaccuracies that do not affect the overall context. \\
        - 4: The explanation is \textbf{entirely factual}, containing only accurate and verifiable information consistent with the image-caption pair. \\

        \midrule
        
        \textbf{Informativeness} \\
        How much relevant detail and meaningful information does the explanation provide to support and justify the score? \\ \\
        
        Scoring Guidelines: \\

        - \textbf{Disagreement Option}: I cannot understand or justify the given score for this image-caption pair. The score is severely misaligned, and I am unable to evaluate the explanation in a meaningful way. \\
        - 1: The explanation is \textbf{not informative at all}, providing little to no relevant information to justify the score, or presenting incorrect or misleading justifications that confuse the evaluator. \\
        - 2: The explanation is \textbf{partially informative}, offering some relevant information but lacking sufficient detail or depth, or providing unclear or slightly inaccurate justifications. \\
        - 3: The explanation is \textbf{fairly informative}, providing useful information but missing important aspects or including some redundant or unnecessary information. \\
        - 4: The explanation is \textbf{highly informative}, offering rich, relevant, and detailed reasoning that thoroughly supports the score. \\

        \bottomrule
    \end{tabularx}
    \caption{Scoring guidelines for consistency, factuality, and informativeness.}
    \label{tab:human_eval_criteria}
\end{center}
\end{table*}
}

\clearpage

\subsection{Evaluation Form}
\label{sec:eval_form}

\begin{figure*}[ht!]
    \centering 
    \includegraphics[width=0.85\linewidth]{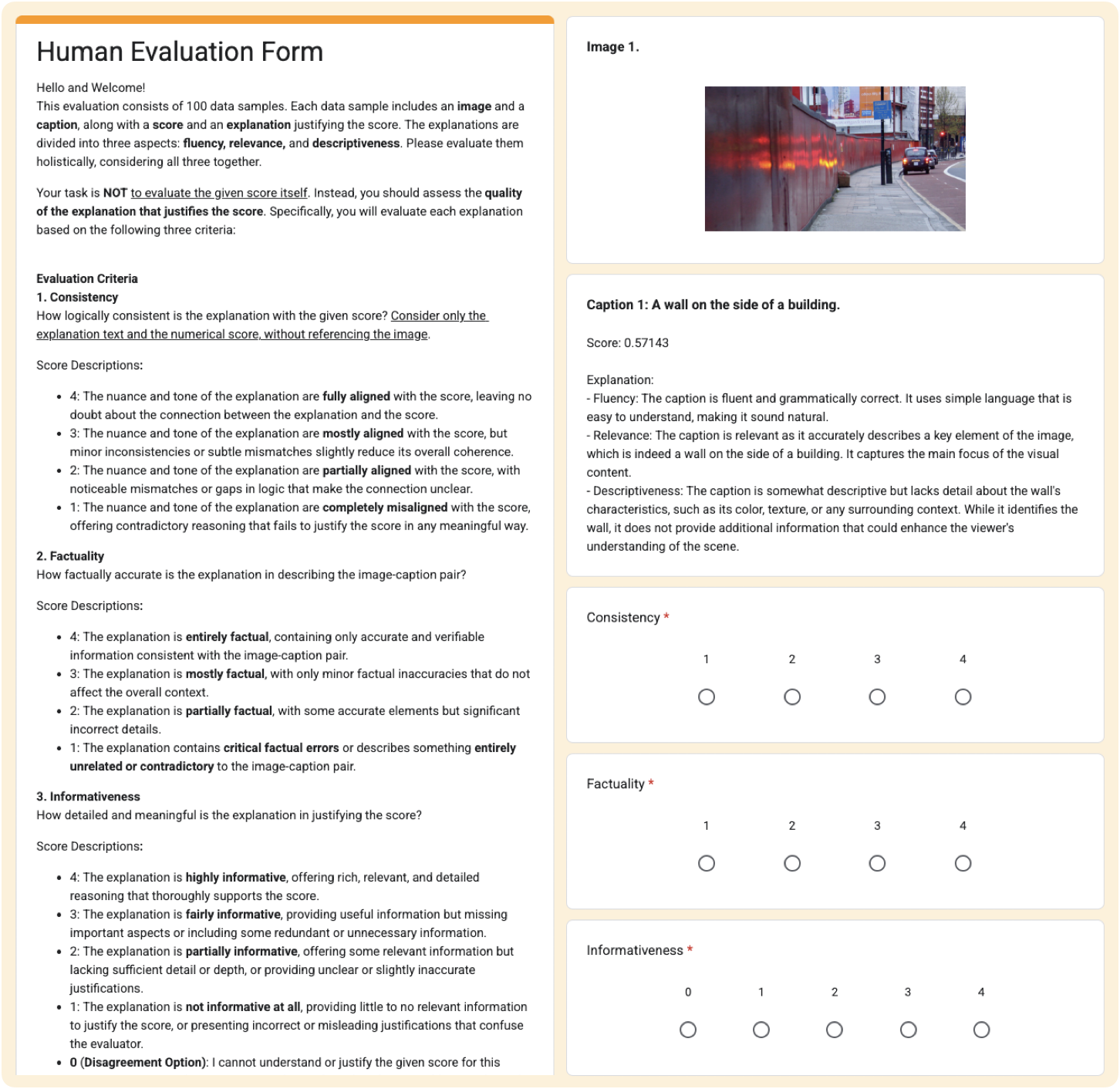}
    \caption{Evaluation form for human evaluation of Polaris-exp and Nebula-exp datasets.} 
    \label{fig:eval_form_gpt}
\end{figure*}

\begin{figure*}[ht!]
    \centering 
    \includegraphics[width=0.85\linewidth]{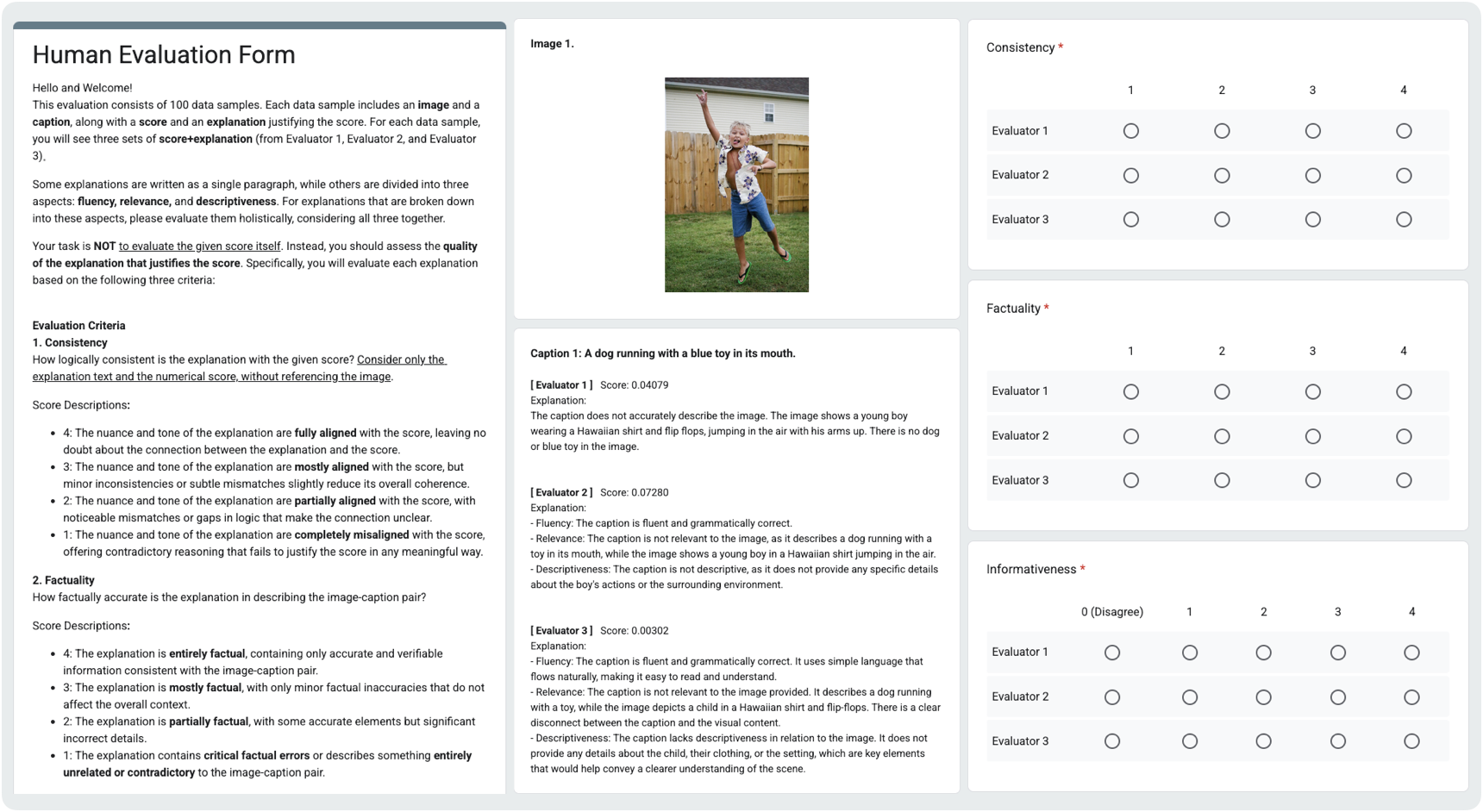}
    \caption{Evaluation form for human evaluation of explanations of FLEUR, $\text{EXPERT}_{\text{w/o SFT}}$, and EXPERT.} 
    \label{fig:eval_form_metrics}
\end{figure*}

\section{Score Binning}
\label{sec:score_binning}

As VLMs recognize each digit of a score as a separate token, not a continuous value as a whole, trivial differences in numerical values (e.g., 0.59375 vs. 0.60) may introduce unnecessary complexity, thereby degrading learning efficiency.
To address this issue, we introduce score binning to guide the model to learn human preferences more effectively.
Human scores in human judgment datasets often include lengthy values due to averaging the scores from multiple annotators. 
Therefore, we intentionally reduce the granularity of these scores by applying score binning.

Table~\ref{tab:score_binning_ablation} presents the evaluation results on benchmark datasets with and without score binning, demonstrating that score binning consistently improves performance across datasets.

\renewcommand{\arraystretch}{1.05}
\begin{table*}[ht!]
\small
\centering
\resizebox{\textwidth}{!}{%
{\small
\begin{tabular}{lccccccccccc}
\toprule
\multirow{2.5}{*}{\textbf{Metric}} & \multicolumn{2}{c}{\textbf{Flickr8k}} & \textbf{COM} & \textbf{Polaris} & \textbf{Nebula} & \multicolumn{5}{c}{\textbf{Pascal-50S} (Accuracy $\uparrow$)} \\
\cmidrule(lr){2-3} \cmidrule(lr){4-4} \cmidrule(lr){5-5} \cmidrule(lr){6-6} \cmidrule(lr){7-11}
 & EX ($\tau_c\uparrow$) & CF ($\tau_b\uparrow$) & ($\tau_c\uparrow$) & ($\tau_c\uparrow$) & ($\tau_c\uparrow$) & HC & HI & HM & MM & Avg \\
\midrule
EXPERT & \textbf{56.7} & \textbf{39.3} & \textbf{65.0} & \textbf{61.1} & \textbf{54.9} & \textbf{62.8} & \textbf{99.8} & \textbf{97.8} & \textbf{78.4} & \textbf{84.7} \\
\phantom{x}w/o Score Binning & 56.2 & 39.1 & \textbf{65.0} & 60.8 & 54.7 & 61.1 & 99.7 & 97.7 & 77.5 & 84.0 \\
\bottomrule
\end{tabular}
}}
\caption{Effect of score binning on performance in human judgment datasets.}
\label{tab:score_binning_ablation}
\end{table*}

\section{LLM-as-a-Judge with GPT-4o}
\label{sec:gpt_discussion}

\subsection{Experimental Setup}
\label{sec:gpt_details}

In our experiments, we include an LLM-as-a-judge baseline using GPT-4o.
We directly prompt GPT-4o with the prompt shown below and the corresponding image, using the \texttt{gpt-4o-2024-08-06} version with the temperature set to 0.
While we closely follow the prompt format used in the scoring stage of EXPERT, we adopt a scale of 0 to 10 instead of 0 to 1 due to output instability. Because of the issues described below, using a scale of 0 to 1 often leads to inappropriate outputs for score smoothing.
We also provide grading criteria for the scores in the prompt, which have been shown to be effective in \citet{lee-etal-2024-fleur}.
\newline

\noindent\texttt{Evaluate the caption and assign a score on a scale of 0 to 10.
\newline
\newline
Grading Criteria: \newline
- 0: The caption does not describe the image at all or is completely incorrect. \newline
- 2: The caption contains minimal accurate details but misses most key elements. \newline
- 4: The caption describes some elements correctly but has significant missing or\\\hspace*{2.3em}incorrect information. \newline
- 6: The caption describes the main elements correctly but lacks some details or\\\hspace*{2.3em}clarity. \newline
- 8: The caption describes the image well with minor omissions or imprecisions. \newline
- 10: The caption accurately and clearly describes all key elements of the image.
\newline
\newline
Caption: \textbf{\{caption\}}
\newline
\newline
Please only provide the score, nothing else.
\newline
Score:
\newline
}

\newpage

\subsection{Experimental Results and Analysis}
\label{sec:gpt_results}

Table~\ref{tab:gpt_discussion_results} presents the experimental results of FLEUR, EXPERT, and GPT-4o \textit{before score smoothing} and \textit{after score smoothing}. Note that the results after score smoothing are the final results reported in Table~\ref{tab:main_results}.

\renewcommand{\arraystretch}{1.05}
\begin{table*}[ht!]
\centering
\resizebox{\textwidth}{!}{%
{\small
\begin{tabular}{clcccccccccc} %
\toprule
  \multirow{2.5}{*}{\textbf{Setting}} & \multirow{2.5}{*}{\textbf{Metric}} & \multicolumn{2}{c}{\textbf{Flickr8k}}  & \textbf{COM} & \textbf{Polaris} & \textbf{Nebula} & \multicolumn{5}{c}{\textbf{Pascal-50S} (Accuracy $\uparrow$ )} \\
\cmidrule(lr){3-4}
\cmidrule(lr){5-5}
\cmidrule(lr){6-6}
\cmidrule(lr){7-7}
\cmidrule(lr){8-12}
&& EX ($\tau_c\uparrow$ ) & CF ($\tau_b\uparrow$ ) & ($\tau_c\uparrow$ ) & ($\tau_c\uparrow$ ) & ($\tau_c\uparrow$ ) & HC & HI & HM & MM & Avg  \\
\midrule
\multirow{3}{*}{\shortstack{Before\\score smoothing}} 
& FLEUR & 34.9 & 51.6 & 58.9 & 53.5 & 52.3 & 26.0 & 99.3 & 92.4 & 42.7 & 65.1 \\
& EXPERT & 35.7 & 52.0 & 58.4 & 54.0 & 57.2 & 26.8 & 99.3 & 93.4 & 45.3 & 66.2 \\
& GPT-4o & \textbf{40.4} & \textbf{54.4} & \textbf{62.9} & \textbf{56.1} & \textbf{61.5} & \textbf{31.5} & \textbf{99.4} & \textbf{95.1} & \textbf{48.0} & \textbf{68.5} \\
\midrule
\multirow{3}{*}{\shortstack{After\\score smoothing}} 
& FLEUR & 53.0 & 38.6 & 63.5 & 58.3 & 51.7 & 61.3 & 99.7 & 97.6 & 74.2 & 83.2 \\
& EXPERT & \textbf{56.7} & \textbf{39.3} & 65.0 & \textbf{61.1} & \textbf{54.9} & \textbf{62.8} & \textbf{99.8} & \textbf{97.8} & \textbf{78.4} & \textbf{84.7} \\
& GPT-4o & 54.3 & \textbf{39.3} & \textbf{65.9} & 58.2 & 54.3 & 60.8 & 99.7 & 97.3 & 73.7 & 82.9 \\
\bottomrule
\end{tabular}
}}
\caption{Experimental results of FLEUR, EXPERT, and GPT-4o before score smoothing and after score smoothing. \textbf{Bold} values indicate the best results for each setting.}
\label{tab:gpt_discussion_results}
\end{table*}

Before score smoothing, GPT-4o achieves the highest performance across all datasets. After score smoothing, however, its performance gains are relatively modest compared to the other models, ultimately leading to inferior results compared to EXPERT.
We attribute this to two main factors:

\begin{itemize}
    \item{\textbf{Tokenizer Differences:}} Since the tokenizer of GPT-4o often outputs multi-digit numbers as a single token, values such as `1', `5', `15', and `02' can all appear as candidate tokens for the same position. This may lead to a more complex and potentially distorted probability distribution for score smoothing, which expects a single-digit number for one token.
    \item{\textbf{Limited Access to Token Probabilities:}} GPT-4o currently provides log probabilities for only the top 20 candidates for each token. We frequently encountered cases where not all digits (0-9) were included, making it impossible to apply score smoothing properly.
\end{itemize}

Due to these constraints, our implementation had to rely only on the available and valid digits, leading to suboptimal performance. Moreover, using closed models like GPT-4o as evaluation metrics involves the following limitations:

\begin{itemize}
    \item{\textbf{Substantial Cost:}} The token-based pricing of proprietary models incurs substantial costs. Table~\ref{tab:gpt_cost} presents the actual expenses from our experiments, where approximately \$106 was spent on the evaluation across all benchmark datasets.
    \item{\textbf{Consistency Over Time:}} Updates to closed models can pose challenges to consistency and reproducibility, both of which are essential qualities for a reliable evaluation metric.
\end{itemize}

In summary, our findings suggest that GPT-4o, despite its strong capabilities across various tasks, does not provide clear benefits over EXPERT as an image captioning evaluation metric.

\renewcommand{\arraystretch}{1.05}
\begin{table*}[ht!]
\centering
\resizebox{\textwidth}{!}{%
{\small
\begin{tabular}{lcccccccccc} %
\toprule
  \multirow{2.5}{*}{\textbf{Metric}} & \multicolumn{2}{c}{\textbf{Flickr8k}}  & \multirow{2.5}{*}{\textbf{COM}} & \multirow{2.5}{*}{\textbf{Polaris}} & \multirow{2.5}{*}{\textbf{Nebula}} & \multicolumn{4}{c}{\textbf{Pascal-50S}} & \multirow{2.5}{*}{\textbf{Total}} \\
\cmidrule(lr){2-3}
\cmidrule(lr){7-10}
& EX & CF & & & & HC & HI & HM & MM &  \\
\midrule
GPT-4o & \$6.64 & \$56.10 & \$15.69 & \$14.05 & \$5.17 & \$2.11 & \$2.11 & \$2.12 & \$2.11 & \textbf{\$106.10} \\
\bottomrule
\end{tabular}
}}
\caption{Summary of GPT-4o evaluation costs across benchmark datasets.}
\label{tab:gpt_cost}
\end{table*}

\clearpage
\twocolumn

\section{Implementation Details}
\label{sec:sft_details}

The configurations for training EXPERT are shown in Table~\ref{tab:config}. 
We apply LoRA \citep{hu2021lora} for SFT.
Two NVIDIA A100 GPUs with 40GB memory were used for SFT, which took approximately 2 hours.

\begin{table}[hbt!]
    \small
    \centering
    \begin{tabular}{lc}
    \toprule
    \textbf{Configuration} & \textbf{Setting} \\
    \midrule
    Epochs & 1 \\
    Batch Size & 8 \\ 
    Optimizer & AdamW \\
    Learning Rate & 2e-05 \\ 
    Weight Decay & 0.0 \\
    Warmup Ratio & 0.03 \\
    Learning Rate Scheduler & Cosine \\
    LoRA Rank & 128 \\
    LoRA Alpha & 256 \\
    \bottomrule
    \end{tabular}
\caption{Training configurations for SFT.}
\label{tab:config}
\end{table}

\section{Inference Time}
\label{sec:inference_time}

Table~\ref{tab:inference_time} presents the inference times for EXPERT and FLEUR on a single NVIDIA A100 GPU.

\begin{table}[hbt!]
    \small
    \centering
    \begin{tabular}{lr}
    \toprule
    \textbf{Metric} & \textbf{Inference Time (sec)} \\
    \midrule
    $\text{FLEUR}_{\text{score}}$ & 0.32 \\ 
    FLEUR & 2.76 \\
    $\text{EXPERT}_{\text{score}}$ & 0.36 \\ 
    EXPERT & 3.80 \\
    \bottomrule
    \end{tabular}
\caption{Inference time of explainable metrics.}
\label{tab:inference_time}
\end{table}

\section{Error Analysis}
\label{sec:error_analysis}

We conducted a detailed error analysis on the 100 samples with the largest absolute difference between the EXPERT score and human score. The errors could be grouped into six main categories:

\begin{table}[hbt!]
    \small
    \centering
    \begin{tabular}{p{0.733\linewidth} r}
    \toprule
    \textbf{Error Type} & \textbf{\# Error} \\
    \midrule
    Overpenalization of Captions Lacking Details & 45 \\
    Overestimation of Captions with Incorrect Details & 17 \\
    Overestimation of Captions Lacking Details & 13 \\
    Misjudgment of Captions with Grammatical Errors & 10 \\
    Human Annotation Errors & 8 \\
    Others & 7 \\
    \midrule
    Total & 100 \\
    \bottomrule
    \end{tabular}
\caption{Categorization of error cases.}
\label{tab:error_analysis}
\end{table}

\newpage

\begin{enumerate}
    \item \textbf{Overpenalization of Captions Lacking \mbox{Details:}} Instances where EXPERT assigned excessively low scores to captions that, while generally accurate, lacked sufficient details.
    \item \textbf{Overestimation of Captions with Incorrect Details:} Instances where EXPERT assigned high scores to captions containing incorrect or misleading content.
    \item \textbf{Overestimation of Captions Lacking \mbox{Details:}} 	Instances where EXPERT assigned high scores to captions that omitted important details.
    \item \textbf{Misjudgment of Captions with Grammatical Errors:} Instances where captions contained grammatical errors, and EXPERT failed to assign appropriate scores.
    \item \textbf{Human Annotation Errors:} Instances where human scores were found to be inaccurate—either higher or lower than what could reasonably be expected.
    \item \textbf{Others:} Errors that do not fall into any of the categories above.
\end{enumerate}

As shown in Table~\ref{tab:error_analysis}, the most common type of error was the overpenalization of captions that lacked sufficient details. A possible solution for this could be to oversample concise captions that received high human scores during training, helping the model better align with human judgment and reduce its bias toward detail-heavy captions.

Furthermore, we analyzed the instances used in the human evaluation in Section~\ref{sec:eval_explanation} that received a score of 2 or below in at least one criterion. Approximately 35\% of these were associated with the \textit{overpenalization of captions lacking details}, where the explanations failed to provide a valid justification for the overpenalization. In the remaining cases, the explanations contained factual inaccuracies regarding elements mentioned in the captions that occupy only a small area in the image, which in turn diminished the informativeness of the explanations. Such difficulty in capturing fine-grained, region-specific details in images is a well-known and fundamental challenge for vision transformers and vision-language models, where active research is underway to address this limitation \citep{guo2024regiongpt, wan2024locca}.

\newpage

\section{Benchmark Datasets}
\label{sec:benchmarks}

\begin{itemize}[leftmargin=1em, itemsep=0pt, topsep=0pt, parsep=2pt, partopsep=0pt]
\item \textbf{Flickr8k} \citep{hodosh2013framing} contains 8,092 images, each paired with approximately five reference captions. The dataset includes two evaluation sets: Flickr8k-Expert (5,664 image-caption pairs evaluated by three experts on a 1-4 scale) and Flickr8k-CF (47,830 pairs assessed by three crowdworkers with binary yes/no judgments, with final scores calculated as the proportion of `yes' responses). Both evaluation sets use 1,000 unique images, with 158 direct reference candidates excluded from the analysis to maintain evaluation integrity.

\item \textbf{COMPOSITE} \citep{aditya2015images} is a multi-source evaluation dataset containing 3,995 images drawn from three collections: MS-COCO (2,007), Flickr8k (997), and Flickr30k (991). It features 11,985 human-evaluated caption pairs, where each image has three candidate captions (one human-authored and two machine-generated) alongside five reference captions. Evaluators rated the caption-image correspondence on a 1-5 scale, from completely irrelevant to perfectly matching. This combination of human and machine-generated captions provides a robust benchmark for comparing automated captioning systems against human performance.

\item \textbf{Polaris} \citep{wada2024polos} represents a significant advancement in image-caption evaluation, incorporating 131,020 human judgments. It features outputs from ten contemporary captioning models, including SAT, M2-Transformer, VinVL, GRIT, BLIPbase, BLIPlarge, GIT, OFA, and two BLIP-2 variants (flan and opt). The dataset draws from both MS-COCO and nocaps collections, ensuring caption diversity. Evaluators rated captions on a five-point scale, considering fluency, relevance, and descriptiveness. The final scores were normalized to [0,1], with careful filtering of unreliable evaluations based on response patterns and timing. Polaris is licensed under the Clear BSD License.

\item \textbf{Nebula} \citep{matsuda2024deneb} expands upon Polaris's foundation, tripling the image count to 32,978. It contains 183,472 reference captions and gathered judgments from 805 evaluators. The dataset's vocabulary spans 32,870 unique words across reference captions (totaling 1,945,956 words) and 3,695 words in candidate captions (288,922 total words). Reference captions average 10.61 words in length, while candidate captions average 8.76 words. To prevent training data leakage, images were sourced from MS-COCO and nocaps validation sets. Evaluators assessed captions using a five-point scale, considering the same criteria as Polaris: fluency, relevance, and descriptiveness.

\item \textbf{Pascal-50S} \citep{vedantam2015cider} draws from the UIUC PASCAL Sentence Dataset \citep{rashtchian-etal-2010-collecting}, comprising 1,000 images—950 with 50 reference captions each and 50 with 120 captions. The dataset features 4,000 evaluation triplets where 48 evaluators compare two candidate captions against the references to determine which better describes the image. The candidate pairs follow four patterns: both correct human-written captions (HC), correct and incorrect human captions (HI), human versus machine-generated captions (HM), and both correct machine-generated captions (MM).
\end{itemize}

\newpage

\section{Baseline Metrics}
\label{sec:baselines}

\begin{itemize}[leftmargin=1em, itemsep=0pt, topsep=0pt, parsep=2pt, partopsep=0pt]
\item \textbf{BLEU} \citep{papineni-etal-2002-bleu} measures translation quality by comparing n-grams between candidate and reference sentences using modified precision with a brevity penalty.
\item \textbf{ROUGE-L} \citep{lin-2004-rouge} evaluates text similarity by finding the Longest Common Subsequence between a candidate and reference sentences, computing recall and precision.
\item \textbf{METEOR} \citep{meteor} evaluates text similarity by aligning words using exact matches, stems, synonyms, and paraphrases, then computes a recall-weighted F-score.
\item \textbf{CIDEr} \citep{vedantam2015cider}  assesses text similarity by using the TF-IDF weighted n-gram similarity between candidate and reference texts, focusing on agreement across multi-references.
\item \textbf{SPICE} \citep{spice} evaluates image captions by converting both candidate and reference captions into scene graphs and comparing their semantic content through scene graph overlap comparison.
\item \textbf{BERTScore} \citep{BERTScore} computes token-level similarity between candidate and reference texts using contextual embeddings from BERT, capturing semantic meaning more effectively than n-gram based metrics.
\item \textbf{CLAIR} \citep{chan-etal-2023-clair} evaluates image captions using LLMs to score and explain caption quality by comparing candidate and reference captions that match human judgments.
\item \textbf{TIGEr} \citep{jiang-etal-2019-tiger} evaluates image captions using a stacked cross-attention network (SCAN), measuring semantic alignment with image regions and similarity to human captions via ranking and grounding distributions.
\item \textbf{ViLBERTScore-F} \citep{lee-etal-2020-vilbertscore} evaluates image captions by incorporating visual context into token embeddings, using ViLBERT to generate image-conditioned representations and compute similarity via cosine similarity.
\item \textbf{CLIPScore} \citep{hessel-etal-2021-clipscore} leverages CLIP embeddings to evaluate image-caption alignment without requiring reference captions, while its variant \textbf{RefCLIPScore} incorporates reference captions by computing the harmonic mean between CLIPScore and reference similarity.
\newpage
\item \textbf{PAC-S} \citep{sarto2023positive} enhances CLIP-based evaluation through positive-augmented contrastive learning using generated image-text pairs, with \textbf{RefPAC-S} incorporating reference captions similar to RefCLIPScore.
\item \textbf{Polos} \citep{wada2024polos} is a multi-modal metric for image captioning that integrates reference-based assessment and learning-based modeling using CLIP and SimCSE-trained RoBERTa.
\item \textbf{FLEUR} \citep{lee-etal-2024-fleur} utilizes LLaVA to directly evaluate captions against images using structured prompts and score smoothing, while \textbf{RefFLEUR} extends this by incorporating reference captions into the evaluation process.
\item \textbf{HICE-S} \citep{zeng2024hicescore} evaluates image captions using global and local image-text similarities, improving hallucination detection. \textbf{RefHICE-S} extends it by adding reference captions via text-text compatibility analysis.
\item \textbf{DENEB} \citep{matsuda2024deneb} evaluates image captions with Sim-Vec Transformer, processing multiple references simultaneously to improve robustness against hallucinations using Sim-Vec Extraction for feature similarity.
\item \textbf{InfoMetIC+} \citep{hu-etal-2023-infometic} is a reference-free metric for image captioning that evaluates captions at both coarse- and fine-grained levels by assessing text precision, vision recall, and token-level alignment with image regions.
\item \textbf{BRIDGE} \citep{sarto2024bridge} employs a dual-encoder architecture with a mapping module that fills masked caption templates using visually enriched pseudo tokens, enabling reference-free caption evaluation.
\end{itemize}

\clearpage
\onecolumn

\section{Additional Examples} %
\label{sec:additional_examples}

\begin{figure*}[ht!]
    \centering 
    \includegraphics[width=1.0\linewidth]{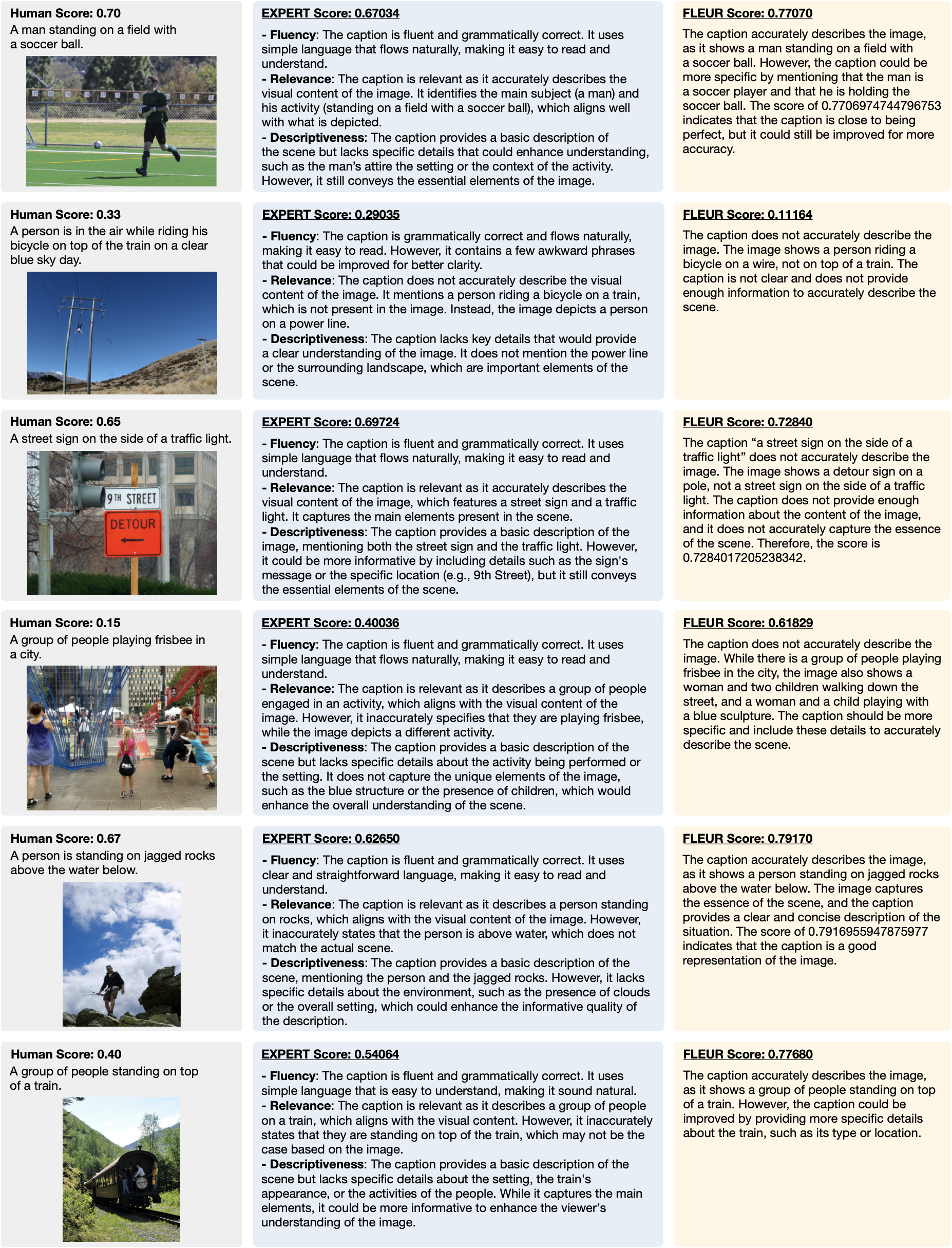}
    \caption{Additional examples of EXPERT.} 
    \label{fig:appendix_examples}
\end{figure*}

\end{document}